\documentclass[10pt,twocolumn,letterpaper]{article}

\usepackage[pagenumbers]{wacv} 

\usepackage{graphicx}
\usepackage{amsmath}
\usepackage{amssymb}
\usepackage{booktabs}
\usepackage{float}
\usepackage{tabularx, array}
\usepackage{multirow}
\usepackage[accsupp]{axessibility}  
\newcolumntype{L}{>{\raggedright\arraybackslash}X}

\usepackage[pagebackref,breaklinks,colorlinks]{hyperref}

\usepackage[capitalize]{cleveref}
\crefname{section}{Sec.}{Secs.}
\Crefname{section}{Section}{Sections}
\Crefname{table}{Table}{Tables}
\crefname{table}{Tab.}{Tabs.}

\def\wacvPaperID{2964} 
\def\confName{WACV}
\def\confYear{2026}

\begin{document}

\title{From Bands to Depth: Understanding Bathymetry Decisions on Sentinel-2}

\author{
Satyaki Roy Chowdhury \quad
Aswathnarayan Radhakrishnan \quad
Hsiao Jou Hsu \quad
Hari Subramoni \\
Joachim Moortgat\\
The Ohio State University, Columbus, OH, USA\\
{\tt\small \{chowdhury.207, radhakrishnan.39, hsu.771,  subramoni.1, moortagat.1\}@osu.edu}
}

\maketitle

\begin{abstract}
Deploying Sentinel-2 satellite derived bathymetry (SDB) robustly across sites remains challenging. We analyze a Swin-Transformer based U-Net model (Swin-BathyUNet) to understand how it infers depth and \emph{when} its predictions are trustworthy. A leave-one-band-out study ranks spectral importance to the different bands consistent with shallow-water optics. We adapt ablation-based CAM to regression (A-CAM-R) and validate the reliability via a performance–retention test: keeping only the top-$p\%$ salient pixels while neutralizing the rest causes large, monotonic RMSE increase, indicating explanations localize on evidence the model relies on. Attention ablations show decoder-conditioned cross-attention on skips is an effective upgrade, improving robustness to glint/foam. Cross-region inference (train on one site, test on another) reveals depth-dependent degradation: MAE rises nearly linearly with depth, and bimodal depth distributions exacerbate mid/deep errors. Practical guidance follows: maintain wide receptive fields, preserve radiometric fidelity in green/blue channels, pre-filter bright high-variance near shore, and pair light target-site fine-tuning with depth-aware calibration to transfer across regions. The source code will be available at \href{https://github.com/OSU-SAI-Lab/From-Bands-to-Depth/tree/main}{https://github.com/OSU-SAI-Lab/From-Bands-to-Depth/tree/main}.

\end{abstract}

\section{Introduction}
\label{sec:intro}
Shallow bays, as well as the water surrounding islands and reefs, concentrate much of humanity’s marine activities, making accurate bathymetric maps of these regions indispensable. As the ocean economy expands—driven by fisheries, oil and gas exploration, and coastal tourism—detailed knowledge of seafloor topography underpins safe navigation, harbor design, and assessments of fishery stocks. Moreover, these shallow-water zones often host critical ecosystems, such as coral reefs and seagrass meadows, whose health and distribution depend on precise depth information; because bathymetric changes reflect broader environmental shifts, including sea‐level rise tied to global warming, monitoring seafloor elevation is also vital for understanding shoreline retreat. Finally, timely bathymetric data play a key role in forecasting and mitigating the impacts of coastal hazards like tsunamis and hurricanes \cite{Kutser2020,Ma2014,Wang2020}.

\subsection{Motivation}
Current shallow-water bathymetric workflows face persistent limitations in scalability, automation, and interpretability. Shipboard sonar (single-beam or multibeam echo sounding) provides high-precision depth data but remains costly and operationally intensive \cite{Kulbacki2024,Bannari2017,Costa2009}. Airborne and spaceborne altimetry extend spatial coverage but lack the resolution required for nearshore or fine-scale mapping. Optical and photogrammetric methods have improved accessibility, yet their accuracy remains tied to water clarity, bottom reflectance, and site-specific tuning.

The advent of hybrid airborne platforms that co-register LiDAR and multispectral imagery now enables the acquisition of large, well-labeled bathymetric datasets more efficiently. However, despite advances in deep learning for satellite-derived bathymetry (SDB), current Sentinel-2 (S2) based models largely emphasize prediction accuracy over interpretability. Few studies investigate what spectral–spatial features actually drive depth inference or how such models generalize across regions. These gaps motivate our development of an interpretable deep framework that fuses active (LiDAR) and passive (optical) sensing to improve both accuracy and transparency in SDB workflows.

\subsection{Challenges}
\begin{enumerate}
  \item \textbf{Coverage–Resolution Imbalance.} Single-beam echo sounding (SBES) offers limited swath width, whereas multibeam echo sounding (MBES) fully insonifies the seabed at higher cost and operational complexity \cite{Kulbacki2024,Bannari2017,Costa2009}.
  \item \textbf{Sensor Limitations.} Airborne LiDAR yields accurate nearshore depths but remains cost-prohibitive at broad scales, while satellite radar altimetry provides only coarse bathymetric detail \cite{Ma2020_ICESat2_S2}.
  \item \textbf{Environmental Constraints.} SAR inversion works only when seabed features match surface wave patterns \cite{Kutser2020}, and optical inversion depends heavily on bottom reflectance and turbidity, limiting transferability \cite{Ji2023,Xie2024,Wu2024}.
  \item \textbf{Photogrammetric Complexity.} Stereo optical reconstructions can recover depths up to $\sim$10\,m but require precise refraction correction and robust image matching \cite{Dietrich2017,DelSavio2023}.
  \item \textbf{Model Interpretability and Transferability.} Few approaches leverage co-registered LiDAR and multispectral data to explore the spectral and spatial cues learned by deep SDB models—leaving both feature attribution and cross-region generalization largely unexplored.
\end{enumerate}

\subsection{Contributions}
This work introduces a practical and interpretable workflow for Sentinel‑2 based satellite derived bathymetry (SDB) and makes the following contributions:  
(1) Swin‑BathyUNet \cite{agrafiotis2025bathymetry} with decoder‑conditioned cross‑attention for robust Sentinel‑2 SDB (Sec.~4.1);  
(2) Leave One Band Out (LOBO) spectral importance analysis consistent with shallow‑water optical physics (Sec.~4.3);  
(3) A‑CAM-R with retention tests for reliable spatial explainability of the model for depth estimation (Sec.~4.4–4.6); and  
(4) Cross‑region, depth‑binned evaluation with actionable guidance for SDB deployment (Sec.~5).

The remainder of this paper is organized as follows: Sec. 3 introduces the MagicBathyNet dataset and its composition across different aquatic environments. Sec. 4 details the Swin‑BathyUNet and its training strategy for bathymetry prediction. Sec. 4 also presents our explainability framework for analyzing spatial and feature‑level model behavior. Sec. 5 reports cross‑region evaluation results, highlighting generalization across unseen domains.

\section{Related Work}
The first spaceborne photon-counting lidar system, ATLAS (Advanced Topographic Laser Altimeter System), is carried by ICESat-2 (Ice, Cloud, and Land Elevation Satellite 2) \cite{Ma2020_ICESat2_S2, Neumann2019, Smith2019}, which was launched in September 2018. It has emerged as a key component of Satellite-Derived Bathymetry (SDB) in recent years as a novel source of a priori bathymetric data that overcomes the drawbacks of traditional satellite-derived techniques that mostly rely on in situ observations. Hsu H. J. et al. \cite{Hsu2021} used ICESat-2 and, S2 images to map shallow depths around six South China Sea islands using a semi-empirical model \cite{Stumpf2003}, whereas Parrish et al. \cite{Parrish2019} used ICESat-2 data to demonstrate 40 m resolution bathymetry in clear waters. Chen Y. et al. \cite{Chen2021} introduced an adaptive variable ellipsoid filtering method (AVEBM) for photon-counting lidar bathymetry and validated its accuracy over Yongle Atoll and the Chilianyu Archipelago. Xie C. et al. \cite{Xie2021} applied DBSCAN clustering to denoise raw ICESat-2 photon returns before fusing them with S2 for bathymetric inversion, showing the promise of multi-source integration. Peng K. et al. \cite{Peng2022} developed PACNN, a physically assisted CNN linking S2 and ICESat-2 inputs for shallow-water depth estimation. Guo X. et al. \cite{Guo2022} used a back-propagation neural network to merge ICESat-2 and S2 data, significantly improving inversion performance. Finally, Xie C. et al. \cite{Xie2024} fused ICESat-2 and S2 data within a radiative-transfer-informed CNN, further enhancing accuracy and underscoring the effectiveness of deep, multi-source fusion.

The majority of multispectral-based depth retrieval techniques use logarithmic band-ratio models to relate reflectance and depth because they take use of the exponential attenuation of light in the water column due to scattering and absorption. Coastal blue, blue, green, and red channels are combined in these methods to lessen vulnerability to changing bottom albedo. Lyzenga \cite{Lyzenga1978} was the first to present this framework, while Stumpf et al. \cite{Stumpf2003} and Lyzenga et al. \cite{Lyzenga2006} further improved it. The Optimal Band Ratio Analysis (OBRA; Legleiter et al. \cite{Legleiter2009}), which methodically chooses the ideal spectral ratios for precise bathymetric inversion, is one of the most popular implementations.

\section{Dataset}

\subsection{MagicBathyNet benchmark}
For learning-based experiments, we adopt \textbf{MagicBathyNet} \cite{agrafiotis2024magicbathynet}, a multimodal benchmark purpose-built for shallow coastal zones. The dataset contains:
\begin{itemize}
  \item \emph{3355} RGB \textbf{triplets} of co-registered patches from S2, SPOT-6, and airborne imagery;
  \item \emph{1244} RGB \textbf{doublets} (S2+SPOT-6);
  \item \emph{Digital Surface Models (DSM)} raster patches (\(3354\) for aerial, \(3396\) for S2/SPOT-6);
  \item \emph{533} annotated raster patches for seabed habitat/type (for pixel-wise classification).
\end{itemize}
Each patch covers \(\mathbf{180\times 180}\,\mathrm{m}\), corresponding to \(18\times 18\) pixels in S2, \(30\times 30\) in SPOT-6, and \(720\times 720\) in airborne imagery. We use MagicBathyNet exactly as released to train/evaluate Swin-Bathy-UNet \cite{agrafiotis2025bathymetry} under standardized splits. MagicBathyNet’s multimodal design (satellite + aerial + DSM + seabed labels) provides both metric depth cues and semantic structure, enabling joint assessment of bathymetry regression.

\begin{figure}[H]
    \centering
    \includegraphics[width=1\linewidth]{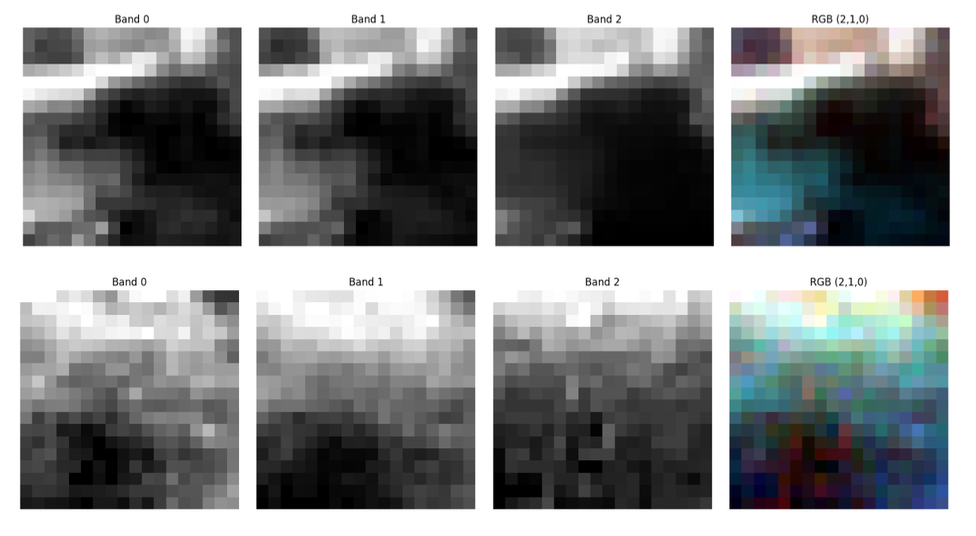}
    \caption{Per-pixel views of Bands 0–2 and their RGB composite (2,1,0) for two example  for Agia Napa region, illustrating shoreline-to-offshore gradients and band-wise noise/contrast differences.}
    \label{fig:image_band}
\end{figure}
\section{Experiments}
\paragraph{Experimental setup (Agia Napa).}
We implemented the Swin-BathyUNet in PyTorch and trained for 30 epochs with Adam optimization \cite{kingma2015adam} and a cosine–annealing learning-rate schedule \cite{loshchilov2017sgdr}, starting at $2.5\times10^{-4}$, on a cluster with 8 NVIDIA L40 GPUs, an AMD EPYC 9124 CPU, and 377~GiB RAM. Supervision used a boundary-weighted RMSE loss with pixel weights linearly scaled in $[1,2]$ to emphasize shoreline transitions. The backbone comprises three ViT blocks (embedding dimensions $\{512,256,128\}$, depth $=1$, $8$ heads, window size $64$, MLP ratio $4$, patch size $32$) feeding a four-stage U-Net with $\{64,128,256,512\}$ filters and dropout rate $0.1$. Inputs are 3-channel RGB patches, and the network predicts a single depth channel.

\begin{figure}
    \centering
    \includegraphics[width=1\linewidth]{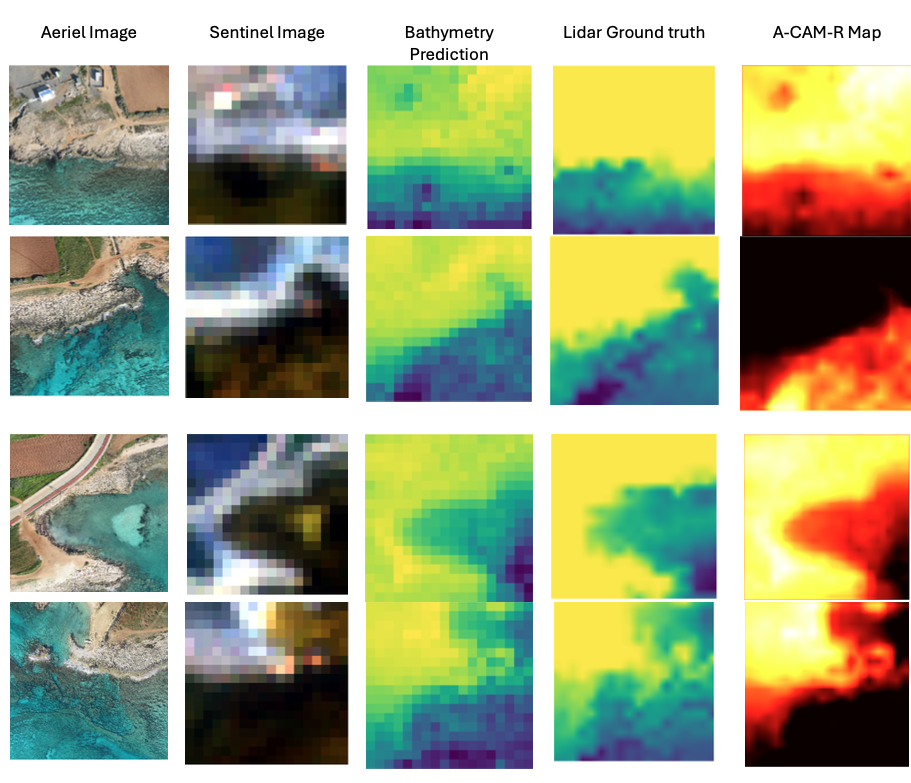}
    \caption{Qualitative results on Agia Napa tiles. \textbf{Left to right}: (i) high-resolution aerial reference, (ii) co-registered S2 chip used by the network, (iii) Swin-BathyUNet bathymetry prediction (meters), (iv) LiDAR ground truth, and (v) A-CAM\textsubscript{R} saliency map. Top to bottom tile number 349, 359, 377, 398 for the Agia Napa region from MagicBathyNet dataset.}
    \label{fig:2}
\end{figure}

\paragraph{Data preprocessing and augmentation.}
RGB orthoimages were normalized to $[0,1]$ by dividing pixel values by $255$, and depths were scaled by $-15$\,m for Agia Napa. All patches were used at their native $720\times720$ resolution. Augmentation included random rotations and random horizontal/vertical flips. Because no near‐infrared band was available, sun–glint mitigation was performed by thresholding overly bright pixels (indicative of specular reflection) and replacing them with the mean of their local neighborhood.

\paragraph{Training/evaluation protocol.}
We trained with paired RGB orthoimage patches and georeferenced, refraction‐corrected SfM–MVS DSM patches. The same RGB patches appear in both training and testing because the prediction targets are DSM gaps that are not visible in the training DSM data. Independent LiDAR and multibeam echo–sounder (MBES) surveys—excluded from both training and testing—served as reference for evaluating those gap areas. Everything was evaluated with the same protocol as \cite{agrafiotis2025bathymetry}.
For \textbf{metrics}, we report root mean square error (RMSE), mean absolute error (MAE) and $R^2$.

\subsection{Attention Ablation Study for Swin-BathyUnet}

\subsubsection{Motivation and Rationale}
Shallow-water bathymetry from multispectral imagery shows long-range dependencies driven by spatially coherent water-column optics (turbidity, sunglint) and slowly varying seabed properties (substrate, ripple fields), which extend beyond local neighborhoods. Purely convolutional U-Net decoders, with their fixed receptive fields, often miss these interactions, causing biased depth transitions and spatially correlated under/over-estimation. To address this, we enrich the skip pathway with windowed self-attention and, when decoder context is present, cross-attention from decoder features to the aligned encoder tokens, under a twofold hypothesis. (1) Windowed self-attention on skip features improves depth continuity and boundary fidelity by aggregating non-local evidence at the same scale. (2) Cross-attention conditions the skip fusion on the decoder’s coarse context (already aware of larger-scale bathymetric structure), prioritizing informative encoder tokens and suppressing distractors (e.g., glint, wakes).

\subsubsection{Ablation Design Focused on Attention}
We evaluate the following variants to isolate attention effects:
\begin{enumerate}
  \item \textbf{Self-only windows at skips}: replace skip $E_\ell$ by $\tilde{E}_\ell=\text{ViT}(E_\ell)$ using windowed self-attention; no cross-attention.
  \item \textbf{Self + Cross at skips}: $\tilde{E}_\ell=\text{ViT}(E_\ell,\text{context}=D_{\ell+1}\!\uparrow)$ using self-attention and cross-attention to decoder context.
\end{enumerate}

\subsection{Supervision design: masks, losses, and training protocol.}

\paragraph{Motivation.}
Bathymetric supervision is spatially sparse and often invalid near shorelines due to occlusions, foams, or SfM/MVS gaps. Penalizing errors uniformly biases training toward large interior regions and away from the difficult transition zones that matter scientifically and operationally (shoreline/breakpoint fidelity). Our experiments therefore (i) restrict the loss to truly valid pixels with a robust mask, and (ii) upweight errors close to the valid/invalid boundary to promote crisp reconstructions near shoreline edges.

\subsubsection*{Target Formation and Normalization}
Let $X\in\mathbb{R}^{C\times H\times W}$ be the S2 stack and $D\in\mathbb{R}^{H\times W}$ the SfM depth (LiDAR is used for evaluation only in this experiment). We normalize the inputs band-wise by reflectance. We simply rescale the S2 bands and depths into a common normalized range and then create a conservative supervision mask that only trains on pixels where both valid depth and valid spectral information are available.
\begin{equation}
\tilde{X}_c(p)\;=\;\mathrm{clip}\!\left(\frac{X_c(p)}{10^{4}},\,0,\,1\right),\qquad c=1,\dots,C, \; p\in\Omega,
\end{equation}
and map depths to $[0,1]$ with a sign-robust scaling:
\begin{align}
\Omega_{\text{valid}} &= \big\{\,p:\; D(p)\neq 0 \ \wedge\ |D(p)|\le 1.5\,D_{\max}\,\big\},\\
s &= -\mathrm{sign}\!\big(\mathrm{median}\{D(p)\,:\,p\in\Omega_{\text{valid}}\}\big),\\
y(p) &= 
\begin{cases}
\mathrm{clip}\!\left(\dfrac{s\,D(p)}{D_{\max}},\,0,\,1\right), & p\in\Omega_{\text{valid}},\\[6pt]
0, & \text{otherwise},
\end{cases}
\end{align}
where $D_{\max}>0$ is a site-specific depth cap ($14.556$\,m for Agia Napa). We construct a conservative supervision mask $M$ by intersecting target and input availability:
\begin{equation}
M(p)\;=\;\mathbf{1}\!\left(y(p)>0\right)\cdot \mathbf{1}\!\left(\frac{1}{C}\sum_{c=1}^{C}\mathbf{1}\!\left(\tilde{X}_c(p)>0\right)>0\right).
\end{equation}

\subsubsection*{Losses (Masked and Boundary-Weighted)}
All objectives are computed on normalized targets $y\in[0,1]$. For reporting in meters we multiply by $D_{\max}$ at the end.

\paragraph{Masked RMSE and MAE.}
\begin{align}
\mathcal{L}_{\text{RMSE}} &= 
\sqrt{\frac{\sum_{p} M(p)\,\big(\hat{y}(p)-y(p)\big)^{2}}{\sum_{p} M(p)+\varepsilon}},\\
\mathcal{L}_{\text{MAE}}  &= 
\frac{\sum_{p} M(p)\,\big|\hat{y}(p)-y(p)\big|}{\sum_{p} M(p)+\varepsilon}.
\end{align}

\paragraph{Boundary-Weighted RMSE (BW-RMSE).}
To emphasize shoreline/validity transitions, errors are upweighted near shoreline/validity boundaries using euclidean distance transform (EDT) so that the model pays more attention to these regions:
\begin{align}
d(p) &= \mathrm{EDT}\big(M\big)(p),\qquad
d'(p)=\mathrm{clip}\!\big(d(p),\,d_{\min},\,d_{\max}\big),
\end{align}
and form a decaying weight (linear or exponential):
\begin{align}
\tilde{w}(p) &=
\begin{cases}
1 - \dfrac{d'(p)-d_{\min}}{d_{\max}-d_{\min}}, & \text{(linear)}\\[8pt]
\exp\!\left(-\dfrac{d'(p)-d_{\min}}{d_{\max}-d_{\min}}\right), & \text{(exp)}
\end{cases}\\
w(p) &= 1 + \frac{\tilde{w}(p)-\min\tilde{w}}{\max\tilde{w}-\min\tilde{w}}\;\in\;[1,2],
\end{align}
yielding the weighted RMSE
\begin{equation}
\mathcal{L}_{\text{BW-RMSE}} \;=\;
\sqrt{\frac{\sum_{p} w(p)\,M(p)\,\big(\hat{y}(p)-y(p)\big)^{2}}{\sum_{p} w(p)\,M(p)+\varepsilon}}.
\end{equation}
Intuitively, $w$ peaks near $M$’s boundary, steering the model to reduce errors where SfM/LiDAR masks transition and optical confounders are prominent.

\subsubsection*{Training and Inference Protocol}
We use $720{\times}720$ crops, batch size $1$, cosine annealing over $E$ epochs, and reflective padding of size $p=16$ at test time to mitigate boundary artifacts:
\begin{equation}
X_{\text{pad}}=\mathrm{pad}_{\text{reflect}}(X,\,p),\qquad 
\hat{y}=\big[\text{Net}(X_{\text{pad}})\big]_{p:H-p,\;p:W-p}.
\end{equation}
We report both the normalized objective and its physical counterpart in meter. Predictions are written to GeoTIFF at the SfM size $(H_t,W_t)$. If the original S2 image had size $(H_s,W_s)$ and affine transform $T_s$, we keep the geospatial extent by rescaling pixel sizes:
\begin{equation}
T_{\text{new}}\;=\;T_s \;\cdot\; \mathrm{diag}\!\left(\frac{W_s}{W_t},\,\frac{H_s}{H_t},\,1\right),
\end{equation}

\subsubsection*{Ablation Plan and Hypotheses:}
We evaluate three objectives under identical data splits and schedules:
\begin{itemize}
    \item \textbf{RMSE} (masked): baseline calibration of squared-error sensitivity.
    \item \textbf{MAE} (masked): robust to outliers; expected to reduce large but sparse mistakes.
    \item \textbf{BW-RMSE}: emphasizes shoreline/boundary neighborhoods.
\end{itemize}

\begin{table}[t]
\centering
\caption{Effect of loss function on Agia Napa S2 image set. Best values in \textbf{bold}. All values in meter.}
\label{tab:loss_ablation_singlecol}
\begin{tabular*}{\columnwidth}{@{\extracolsep{\fill}}lccc@{}}
\toprule
Loss & RMSE$\downarrow$ & MAE$\downarrow$ & $R^2\uparrow$ \\
\midrule
RMSE      & 1.353 & 0.919 & 0.9314 \\
MAE       & \textbf{1.243} & \textbf{0.817} & \textbf{0.9421} \\
BW-RMSE   & 1.305 & 0.856 & 0.9362 \\
\bottomrule
\end{tabular*}
\end{table}

\subsection{Band-Importance via LOBO}
We assess each S2 band’s importance for depth estimation via a leave-one-band-out ablation: holding the trained network fixed, we remove a single band and record the resulting increase in error. To quantify spectral contributions, we perform \emph{leave-one-band-out} (LOBO) on the trained network. Let $x\in[0,1]^{C\times H\times W}$ be the S2 stack and $x^{(-b)}$ be $x$ with band $b$ replaced by a neutral fill $\phi_b$ (we use the mean over valid pixels):
\begin{align}
x^{(-b)}_c = 
\begin{cases}
x_c, & c\neq b,\\
\phi_b\mathbf{1}, & c=b.
\end{cases}
\end{align}
For each test tile, we obtain $\hat{y}_\text{full}=f(x)$ and $\hat{y}_{(-b)}=f(x^{(-b)})$, compute RMSEs against the normalized ground truth on valid pixels, and aggregate:
\begin{align}
\Delta\mathrm{RMSE}_b \;=\; \mathrm{RMSE}\big(\hat{y}_{(-b)}\big) \;-\; \mathrm{RMSE}\big(\hat{y}_\text{full}\big),
\end{align}
reported in meters after de-normalization. Positive $\Delta\mathrm{RMSE}_b$ indicates band $b$ carries useful information; magnitude reflects its importance. This procedure controls for architecture and training stochasticity by reusing a \emph{fixed} network and perturbing inputs only. $\Delta\mathrm{RMSE}_b$ is conditional on the trained model and site; importance can shift with training loss and domain. Correlated bands may mask each other's effect and LOBO captures marginal rather than joint contributions.

\begin{table}[t]
\centering
\caption{LOBO band importance. $\Delta\mathrm{RMSE}_b$ is the error increase (m) when band $b$ is replaced by its masked mean. Higher is more important.}
\label{tab:lobo}
\begin{tabular*}{\columnwidth}{@{\extracolsep{\fill}}lccc@{}}
\toprule
Band & $\Delta\mathrm{RMSE}_b$ (m)$\uparrow$ & Rank \\
\midrule
B1 (Green)      & 0.818 & 1 \\
B0 (Blue)       & 0.531 & 2 \\
B2 (Red)        & 0.223 & 3 \\
\bottomrule
\end{tabular*}
\end{table}

\subsection{Visual explanations with A‑CAM‑R}
\label{subsec:acamr_motivation}
Bathymetry from S2 is governed by subtle, spatially distributed cues—color shifts over seabed types, radiance gradients near the  shore and breaker patterns. While aggregate metrics (RMSE/MAE) quantify \emph{how well} a model predicts depth, they do not reveal \emph{where} the evidence originates, whether the model relies on physically meaningful regions, or when errors stem from artifacts (e.g., sunglint, seams, boats and vegetation). This limits trust, weakens cross-site claims, and provides little guidance for data curation or architecture choices.

To bridge this gap, we adapt Ablation-CAM \cite{desai2020ablationcam} to Sentinel depth estimation (\textbf{A-CAM-R}) and turn interpretability into a \emph{measurable} property via a performance-retention test. Let $E_{\text{full}}$ denote the baseline error (in meters) computed on valid water pixels. From the last decoder block activations $\{A_k\}_{k=1}^K$, we ablate each channel to obtain an error $E_{-k}$ and define an importance weight
\[
w_k \;=\; \frac{E_{-k}-E_{\text{full}}}{E_{\text{full}}+\varepsilon}.
\]
The localization map is the ReLU-weighted sum
\[
L_{\text{A-CAM-R}} \;=\; \mathrm{ReLU}\!\Big(\sum_{k=1}^{K} w_k\,A_k\Big),
\]
upsampled to input resolution.

\begin{table}[t]
\centering
\caption{Performance–retention on selected Agia Napa tiles using A\textendash CAM\textsubscript{R}. 
$E_{\text{full}}$ is the baseline RMSE (m). For each kept area $p\%$, we report the masked-input error $E_{\text{mask}}$ (m) and the relative increase $\Delta$ (lower is better).}
\label{tab:acamr_retention_selected}
\scriptsize
\setlength{\tabcolsep}{3.5pt}
\begin{tabular*}{\columnwidth}{@{\extracolsep{\fill}}l c c c c@{}}
\toprule
\multirow{2}{*}{Tile} &
\multirow{2}{*}{$E_{\text{full}}$ (m)} &
\multicolumn{3}{c}{$E_{\text{mask}}$ (m) / $\Delta$ (\%)} \\
\cmidrule(lr){3-5}
& & 20\% & 30\% & 50\% \\
\midrule
349 & 0.237 & 1.067 / 350.5 & 1.043 / 340.4 & 0.973 / 310.7 \\
359 & 0.422 & 1.571 / 272.3 & 1.442 / 241.7 & 0.826 / 95.8 \\
377 & 0.275 & 1.041 / 278.9 & 1.019 / 270.8 & 0.978 / 256.1 \\
398 & 0.450 & 2.113 / 369.4 & 1.981 / 340.0 & 1.842 / 309.2 \\
\bottomrule
\end{tabular*}
\vspace{2pt}
\end{table}

\begin{table}[t]
\centering
\caption{Ablation on attention  on Agia Napa S2 images. Best numbers in \textbf{bold}. All values in meter}
\resizebox{\columnwidth}{!}{%
\begin{tabular}{lcccccc}
\toprule
Variant & RMSE$\downarrow$ & MAE$\downarrow$ & $R^2\uparrow$ \\
\midrule
A1: Cross-only (Swin) & \textbf{1.262} & \textbf{0.840} & \textbf{0.9404} &  \\
A2: Self-only         & 1.365 & 0.907 & 0.9302 &  \\
A3: Self+Cross        & 1.275 & 0.865 & 0.9391 &  \\
\bottomrule
\end{tabular}%
}
\label{tab:attn_ablation}
\end{table}

\subsection{Notation and data}
Let $\mathbf{x}\in[0,1]^{C\times H\times W}$ be a Sentinel–2 tile rescaled to $[0,1]$ per band. Unless noted, bands are Blue ($B0$), Green ($B1$), Red ($B2$). The model predicts a normalized depth map $\hat{\mathbf{y}}\in[0,1]^{H\times W}$; ground-truth (LiDAR) normalized depth is $\mathbf{y}\in[0,1]^{H\times W}$ with valid‐water mask $\mathcal{M}\subset\{1,\dots,H\}\times\{1,\dots,W\}$. Denormalization uses the site-specific scale $D_{\max}>0$ (meters), so physical units are $\hat{\mathbf{y}}D_{\max}$ and $\mathbf{y}D_{\max}$.
We report RMSE in meters over valid pixels, the following steps measure how much the depth error increases when only the top‑salient pixels are kept and the rest are neutralized. 

For a retention level $p\in\{20,30,50\}$ (percent), let $\Omega_p\subset\{1,\dots,H\}\times\{1,\dots,W\}$ be the top-$p\%$ pixels of $L$. We construct a \emph{neutralized} input $\tilde{\mathbf{x}}^{(p)}$ by keeping original values inside $\Omega_p$ and filling all bands outside $\Omega_p$ by a per-band neutral value (the mean of that band over $\Omega_p\cap\mathcal{M}$ if available, else the image mean). We re-infer $\hat{\mathbf{y}}^{(p)}$ on $\tilde{\mathbf{x}}^{(p)}$ and compute:
\[
\begin{aligned}
E_{\mathrm{mask}}(p) &= \mathrm{RMSE}\!\big(\hat{\mathbf{y}}^{(p)},\,\mathbf{y};\,\mathcal{M}\big),\\
\Delta(p) &= 100\,\frac{E_{\mathrm{mask}}(p)-E_{\mathrm{full}}}{E_{\mathrm{full}}}\,\%.
\end{aligned}
\]

A reliable explanation yields \emph{monotonic} degradation: smaller $p$ (less evidence kept) $\Rightarrow$ larger $E_{\mathrm{mask}}(p)$ and positive $\Delta(p)$.

\subsection{Feature Correlation with Saliency Map}
Inside $\Omega_p$ we compute interpretable 2D feature maps, each aligned to the image grid:
\begin{itemize}
\item \textbf{VIS brightness}: $\mathrm{VIS}=\tfrac{1}{3}(B0+B1+B2)\in[0,1]$.
\item \textbf{Band ratios}: 
$\mathrm{R}_{0/1}=\frac{B0}{B1+\epsilon}$ and 
$\mathrm{R}_{1/2}=\frac{B1}{B2+\epsilon}$, clipped to a finite range.
\item \textbf{Local variance}: computes how much the VIS values fluctuate within a small neighborhood by measuring their variance

$\mathrm{Var}_{\ell}=\mathbb{E}[\mathrm{VIS}^2]-\big(\mathbb{E}[\mathrm{VIS}]\big)^2$ over a $k\times k$ window (we use $k{=}7$).
\item \textbf{Gradient energy} (edges/structure): $\mathrm{Grad}=\|\nabla\mathrm{VIS}\|$ via a Sobel operator.
\item \textbf{Fine texture (Laplacian variance)}:
Let $S = G_{\sigma} * \mathrm{VIS}$ be a Gaussian-smoothed VIS (e.g., $\sigma\!\approx\!1$), and $L=\nabla^2 S$ its Laplacian. 
Define the local mean $\mu_L^{(k)}$ over a $k\times k$ window and
\[
\mathrm{LapVar}_k=\mathbb{E}_{k}[L^2]-\big(\mathbb{E}_{k}[L]\big)^2,
\]
which increases with fine-scale texture.

\end{itemize}
For each feature $F\in\{\mathrm{VIS},\mathrm{Var}_{\ell},\mathrm{Grad},\mathrm{R}_{0/1},\mathrm{R}_{1/2},\dots\}$ we summarize over $\Omega_p$ its mean, 90th percentile, and the Pearson-style correlation with saliency:
\[
\mathrm{corr}_s(F,L)=\tfrac{1}{|\Omega_p|}\sum_{(i,j)\in\Omega_p}
\frac{F_{ij}-\mu_F}{\sigma_F+\epsilon}\cdot
\frac{L_{ij}-\mu_L}{\sigma_L+\epsilon}\in[-1,1].
\]
Positive correlation means the model’s saliency aligns with that cue (relied upon); negative means it is down-weighted or avoided.

High-resolution bathymetry from passive imagery is limited by (i) scarce, costly labels (site-specific SfM/LiDAR) and (ii) the need to map new areas quickly without ground truth data. A model trained once on a well-surveyed region and transferable with little or no relabeling would deliver immediate value for coastal monitoring, disaster response, and baseline mapping. Cross-region inference is a principled test of \emph{generalization under domain shift}. Site-to-site changes in water optics (turbidity, chlorophyll), atmosphere, sun–sensor geometry, seabed, and depth distributions alter bottom-to-sensor radiance and can break the shortcuts learned during training. Evaluating a fixed model on a new region allows us to (i) quantify error growth with depth and sampling density, (ii) disentangle \emph{data imbalance} from physics-driven limits, and (iii) identify spectral/geometric cues that transfer reliably. Cross-region inference is essential as it reveals failure modes, defines a safe operating window, and highlights cost-effective steps, fine-tuning, depth-aware calibration and domain-robust augmentations for practical and scalable SDB.

\begin{table*}[t]
\centering
\small
\setlength{\tabcolsep}{6pt}
\renewcommand{\arraystretch}{1.08}
\caption{\textbf{Feature–aware summary for representative tiles (sign of correlation with saliency).} “Ratios” abbreviate $B0/B1$ and $B1/B2$. Grad = gradient energy on VIS; Var = local variance. Arrows indicate positive (\(\uparrow\)), negative (\(\downarrow\)), or near-zero (\(\approx 0\)) correlation inside A–CAM–R region.}
\label{tab:acamr_tiles}
\begin{tabularx}{\textwidth}{@{}c c c c c >{\raggedright\arraybackslash}X@{}}
\toprule
\textbf{Tile} & \textbf{Ratios} & \textbf{VIS} & \textbf{Grad} & \textbf{Var} & \textbf{Values (correlations)} \\
\midrule
349 & \(\downarrow\) (both) & \(\downarrow\) & \(\downarrow\) & \(\downarrow\) &
R$_{0/1}=-0.747$, R$_{1/2}=-0.897$; VIS $=-0.148$; Grad $=-0.417$; Var $=-0.278$; LapVar $=-0.00187$ \\
\addlinespace[2pt]
359 & \(\uparrow\) (both) & \(\approx 0\) & \(\downarrow\) & \(\downarrow\) &
R$_{0/1}=+0.794$, R$_{1/2}=+0.523$; VIS $=+0.085$; Grad $=-0.131$; Var $=-0.083$; LapVar $=-0.00029$ \\
\addlinespace[2pt]
377 & \(\downarrow\) (both) & \(\approx 0\) & \(\downarrow\) & \(\downarrow\) &
R$_{0/1}=-0.274$, R$_{1/2}=-0.514$; VIS $=+0.022$; Grad $=-0.117$; Var $=-0.092$; LapVar $=-0.00127$ \\
\addlinespace[2pt]
398 & \(\downarrow\) (both) & \(\uparrow\) & \(\approx 0\) & \(\uparrow\) (fine) &
R$_{0/1}=-0.555$, R$_{1/2}=-0.837$; VIS $=+0.598$; Grad $=-0.032$; Var $=-0.020$; \textbf{LapVar} $=\mathbf{+0.00145}$ \\
\bottomrule
\end{tabularx}
\end{table*}

\section{Results}
\label{sec:results}

We report results from four complementary lenses: (i) supervision design and masking, (ii) architectural attention ablations, (iii) spectral importance via LOBO, and (iv) spatial reliability via A–CAM-R with performance–retention. Unless stated otherwise, all errors are reported in meters on valid water pixels.

\vspace{0.25em}
\noindent\textbf{Effect of different loss function:}
Table~\ref{tab:loss_ablation_singlecol} compares masked RMSE, masked MAE, and the boundary-weighted RMSE (BW-RMSE). Masked MAE attains the best overall accuracy (\textbf{RMSE}=$1.243$, \textbf{MAE}=$0.817$, $R^2=0.9421$), indicating that a robust \(\ell_1\) penalty better tolerates sparse, large residuals common in shallow zones and at instrument seams than \(\ell_2\). BW-RMSE (\textbf{RMSE}=$1.305$, \textbf{MAE}=$0.856$, $R^2=0.9362$) trades a small amount of bulk accuracy for improved error distribution. Masked MAE yields the best overall accuracy, while BW‑RMSE slightly sacrifices bulk error to improve shoreline fidelity.

\vspace{0.25em}
\noindent\textbf{Attention ablations:}
Table~\ref{tab:attn_ablation} shows that \emph{Cross-only} yields the best aggregate performance (\textbf{RMSE}=$1.262$, \textbf{MAE}=$0.840$, $R^2=0.9404$), indicating that decoder-conditioned cross-attention on skips effectively suppresses distractors and captures long-range structure. \emph{Self-only} performs worst (RMSE=$1.365$), suggesting that non-contextual mixing amplifies nuisances, while \emph{Self+Cross} (RMSE=$1.275$, MAE=$0.865$) recovers most \emph{Cross-only} gains but does not surpass it. Overall, retaining large effective receptive fields and prioritizing \emph{context-aware} skip fusion via decoder-conditioned cross-attention provides the best accuracy–efficiency tradeoff.

\vspace{0.25em}
\noindent\textbf{Spectral importance (LOBO):}
Table~\ref{tab:lobo} ranks S2 bands by marginal contribution: \(\text{B1/Green}\) is most informative (\(\Delta\mathrm{RMSE}=0.818\) m), followed by \(\text{B0/Blue}\) (\(0.531\) m) and \(\text{B2/Red}\) (\(0.223\) m). This ordering matches shallow-water optics (green penetrates deepest, blue aids very shallow/clear, red attenuates rapidly). Practically: (i) preserve high-quality green, (ii) mitigate green–blue artifacts (e.g., glint), and (iii) avoid over-allocating capacity to red unless turbidity/substrate warrant it. As LOBO captures \emph{marginal} effects on a fixed model, importance can shift with site, loss, and augmentation, but the ranking is physics-consistent and actionable for data curation and sensor selection.

We validate A–CAMR by keeping only the top-\(p\)\% salient pixels and neutralizing the rest of the S2 input (Table~\ref{tab:acamr_retention_selected}). Across representative tiles (349/359/377/398), masking outside the top regions inflates error substantially. For instance, at \(p{=}20\%\), RMSE rises from \(0.237\to1.067\) m (+350\%) on tile 349 and \(0.450\to2.113\) m (+369\%) on tile 398; even at \(p{=}50\%\), errors remain far above baseline (e.g., 349: \(0.237\to0.973\) m; 398: \(0.450\to1.842\) m). These large, monotonic degradations indicate that A–CAMR localizes pixels the model \emph{truly} relies on; they also reveal that Swin-BathyUNet aggregates evidence over broad spatial support (shoreline gradients, bars, substrate transitions), not just small hotspots. This justifies preserving wide context during preprocessing (crop sizes, padding) and favors designs with strong multi-scale paths. Conversely, tiles that show saliency concentrated on obvious artifacts can be flagged for filtering or targeted augmentation (e.g., glint suppression), turning interpretability into a practical quality-control signal.

Figure~\ref{fig:2} (Agia Napa tiles 349, 359, 377, 398) illustrates three consistent patterns: (i) predicted depth maps track LiDAR morphology with improved shoreline/breakpoint fidelity under BW-RMSE supervision; (ii) A–CAM-R heatmaps emphasize optically shallow structures and substrate boundaries while de-emphasizing uniform deep water and (iii) failure modes align with optical confounders (glint, foam, wakes), exactly where cross-attention helps by injecting decoder context. These observations align with the quantitative trends above and provide concrete hooks for dataset triage (artifact filtering) and for receptive-field tuning (context-aware skips). 

From Table ~\ref{tab:acamr_tiles}, it is evident that clear optics (e.g., tile 359), $\mathrm{R}_{0/1}$ and $\mathrm{R}_{1/2}$ correlate positively with saliency, while edge/texture cues are downweighted: the network leverages blue/green penetration consistent with underwater optics. Under turbidity or mixed substrate (tiles 349, 377), both ratios and edge/texture cues correlate negatively; the model instead aggregates broader spatial context via decoder-conditioned cross-attention. Tile 398 shows positive correlation with VIS and fine texture but negative correlation with ratios and slope, a signature of breakers/foam or sun-glint near shore. A lightweight quality-control (QC) gate that suppresses pixels with high VIS and high $\mathrm{Var}_{\ell}$ close to shore mitigates this failure mode. Binary edge maps contribute little; continuous gradient energy often correlates negatively, indicating the model avoids sharp visual edges (frequently artifacts) and prefers multi-scale, spatially coherent cues—precisely what context-aware skip fusion encourages.
Across tiles, $E_{\mathrm{mask}}(p)$ increases as $p$ decreases (top-20\% $>$ 30\% $>$ 50\%), confirming that A–CAM–R localizes \emph{causal} evidence rather than incidental structure. High-saliency regions are spatially \emph{extended}, indicating the model integrates information over broad support (shoreline gradients, bars, substrate transitions), which supports the use of windowed attention and cross-attention. From the experiment in Fig. ~\ref{fig:depth bin} we can see that:

\begin{figure}
    \centering
    \includegraphics[width=1\linewidth]{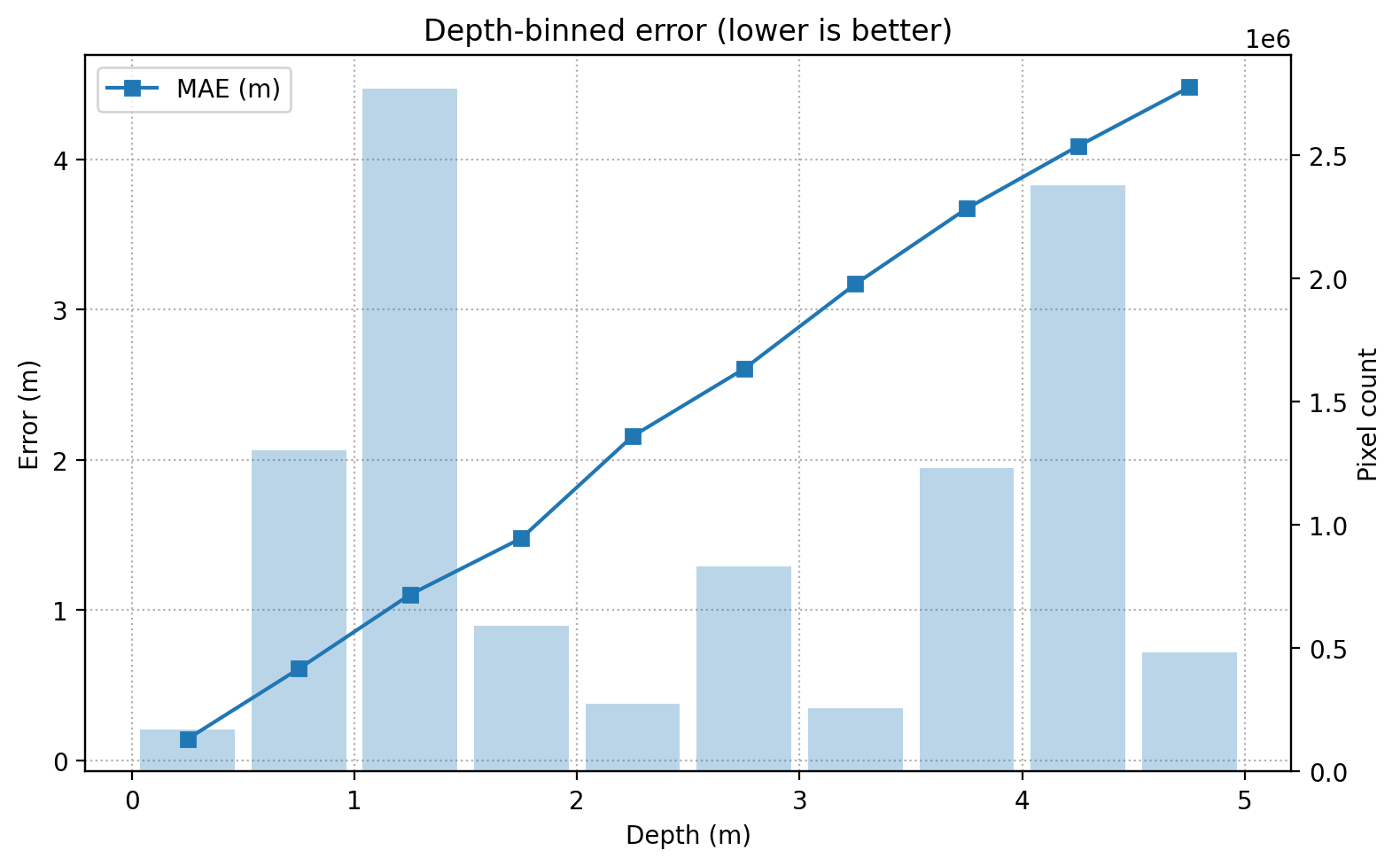}
    \caption{Depth-binned bathymetry error (lower is better). MAE is plotted against depth for the first 20 images of Puck Lagoon area in the MagicBathyNet dataset using the model trained on Agia Napa region; light-blue bars denote pixel counts on the secondary axis. Errors increase nearly linearly with depth, while a bimodal pixel distribution suggests data imbalance contributes to degradation at depth.}
    \label{fig:depth bin}
\end{figure}

\begin{itemize}
  \item \textbf{MAE vs.\ depth:} The orange MAE curve increases nearly linearly with depth, from $\approx0.2$\,m at $\sim0.25$\,m to $\approx4.5$\,m by $\sim4.75$\,m. This indicates a rapid loss of predictive skill as water gets deeper.
\end{itemize}

\begin{itemize}
  \item \textbf{Pixel-count distribution:} The light-blue bars (right axis) show a bimodal depth distribution: many pixels in very shallow water ($\approx0.7$--$1.5$\,m) and another concentration near $\approx4.2$\,m, with fewer samples in the $2$--$3.5$\,m range.
  \item \textbf{Implication:} This imbalance likely biases learning toward shallow patterns and leaves the model under-constrained in mid/deeper water, contributing to degraded performance with increasing depth.
\end{itemize}

\begin{itemize}
  \item The model is \textbf{most reliable} in very shallow water ($<\!\sim1$\,m) and \textbf{degrades steadily} with depth; beyond $\sim3$\,m the absolute errors are often operationally prohibitive for this site/configuration.
  \item Missing RMSE at depth reflects \textbf{data scarcity or masking}, not necessarily improved performance.
\end{itemize}

The model was trained on a different geographic region, so the evaluation site introduces a \emph{domain shift} in optics and seafloor conditions. Several observations support that the depth-binned errors in Figure ~\ref{fig:depth bin} are consistent with cross-region mismatch:

\paragraph{Physical/covariate differences.}
Water-leaving radiance and visibility at depth depend on turbidity, chlorophyll, CDOM, aerosols, sun–sensor geometry, and bottom albedo. If these factors differ from the training region, the learned mapping $(\text{bands}\!\rightarrow\!\text{depth})$ becomes biased—especially where water-column attenuation is strongest (larger depths). The near-linear growth of MAE with depth (Fig.~X) is consistent with greater attenuation and altered bottom reflectance spectra relative to training.

\paragraph{Data imbalance interaction.}
The evaluation set is bimodal in depth (many very shallow pixels and a cluster near $\sim4$–$5$\,m), while training depth statistics differ (not shown). Such distributional mismatch amplifies error where the training set had fewer examples (mid/deep bins).

\section{Conclusion}
In this study, we aimed to pair Swin‑BathyUnet with model‑aware analysis to convert SDB into an interpretable and trustworthy workflow. \textbf{What we demonstrated:} A‑CAM‑R with retention masks confirms reliableness—errors rise as evidence is removed—while diagnostics reveal when the model leans on green/blue cues in clear shallows versus a broader context under turbidity. Decoder‑conditioned cross‑attention is the most cost‑effective upgrade; masked MAE optimizes the overall accuracy, and a boundary‑weighted RMSE sharpens the shorelines. Practically, wide receptive fields are preserved, radiometric fidelity in B0/B1 is prioritized, and bright, high‑variance near‑shore glint/foam is gated out. Cross‑region tests expose depth‑dependent degradation under domain shift, which can be mitigated by restricting use to shallow water, applying light target‑site fine‑tuning, using physics‑informed learning, and deploying depth‑aware calibration with uncertainty plus robust augmentations. \textbf{Future work:} Multi‑site training, explicit uncertainty calibration, temporal robustness, and physics‑assisted fusion (e.g., with ICESat‑2) are expected to further enhance transferability and reliability.

\section{Acknowledgements}
We would like to thank Dr. Joachim Moortgat and Hsiao Jou Hsu for their constant support and guidance. This research is supported by Dr. Subramoni's faculty development and Dr. Moortgat's and Dr. Subramoni's presidential research excellence grant from The Ohio State University. The authors are grateful for the generous
support of the computational resources from the Arts and Sciences Technology Services (ASCTech).


\bibliographystyle{ieee_fullname}
\bibliography{refs}

\end{document}